\documentclass[conference]{IEEEtran}
\IEEEoverridecommandlockouts
\usepackage{tcolorbox}
\usepackage{multirow}
\usepackage{mathtools}
\usepackage{enumitem}
\usepackage{cite}
\usepackage{amsmath,amssymb,amsfonts}
\usepackage{algorithmic}
\usepackage{graphicx}
\usepackage{textcomp}
\usepackage{xcolor}
\usepackage{hyperref}
\usepackage{caption}

\def\BibTeX{{\rm B\kern-.05em{\sc i\kern-.025em b}\kern-.08em
    T\kern-.1667em\lower.7ex\hbox{E}\kern-.125emX}}

\begin{document}

\title{Deep Fraud Detection on Non-attributed Graph}

\author{\IEEEauthorblockN{Chen Wang\IEEEauthorrefmark{1}\IEEEauthorrefmark{3}\thanks{\IEEEauthorrefmark{3}The work was done during Chen Wang's internship at Grab.},
Yingtong Dou\IEEEauthorrefmark{1},
Min Chen\IEEEauthorrefmark{2}, 
Jia Chen\IEEEauthorrefmark{2},
Zhiwei Liu\IEEEauthorrefmark{1} and Philip S. Yu\IEEEauthorrefmark{1}}
\IEEEauthorblockA{\IEEEauthorrefmark{1}Department of Computer Science, University of Illinois at Chicago, USA}
\IEEEauthorblockA{\IEEEauthorrefmark{2}GrabTaxi Holdings Pte Ltd, Singapore\\ \{cwang266, ydou5, zliu213, psyu\}@uic.edu, \{min.chen, jia.chen\}@grab.com}}

\maketitle

\begin{abstract}
Fraud detection problems are usually formulated as a machine learning problem on a \textit{graph}.
Recently, Graph Neural Networks (GNNs) have shown solid performance on fraud detection.
The successes of most previous methods heavily rely on rich node features and high-fidelity labels.
However, labeled data is scarce in large-scale industrial problems, especially for fraud detection where new patterns emerge from time to time.
Meanwhile, node features are also limited due to privacy and other constraints.
In this paper, two improvements are proposed:
1) We design a graph transformation method capturing the structural information to facilitate GNNs on non-attributed fraud graphs.
2) We propose a novel graph pre-training strategy to leverage more unlabeled data via contrastive learning.
Experiments on a large-scale industrial dataset demonstrate the effectiveness of the proposed framework for fraud detection.
\end{abstract}


\section{Introduction}
\label{sec01-intro}
The rapid growth of online services facilitates people's life while fosters fraudsters who reap monetary rewards and users' privacy via tampering with the system and policy.
To name a few, the review spammer could boost the reputation of dishonest merchants in e-commerce and sabotage the recommender system~\cite{zhang2020gcn}.
Meanwhile, an increasing number of fraudsters have been engaging in social platforms and online financial services, according to a recent report~\cite{datavisor}.

To combat the fraudsters automatically, many machine learning approaches have been proposed~\cite{jiang2016suspicious}.
As Graph Neural Networks (GNNs) achieving superior performance on many graph-related tasks~\cite{kipf2016semi, hamilton2017inductive},
many researchers and practitioners begin to adopt GNNs to detect fraud in various scenarios~\cite{liu2018heterogeneous, li2019spam, dou2020enhancing, liu2021intention}.
Most GNNs hold the graph homophily assumption where the connected nodes in a graph should have similar properties.
Specifically, GNN recursively aggregates the node and its neighbor information to learn the node representation.

For the fraud detection problem, fraudsters usually behave insidiously, but their suspicious signals can be magnified when connecting them via shared entities like the IP address and the device~\cite{liu2018heterogeneous}.
For instance, if a group of coordinated fraudsters frequently use the same IP address and device, they would be closely connected on graphs built with the above entities.
On the contrary, benign entities are more independent on the graph since they do not have coordinated behavior. 
After aggregating the neighbor information of the entities in the above graph using GNNs, fraudster entities' suspiciousness will become more significant compared to benign ones.

Despite the effectiveness of GNN-based fraud detectors, most of them depend on highly personalized graphs coupled with corresponding data which are not applicable to other problems~\cite{liu2018heterogeneous, liu2021intention}.
Meanwhile, the majority of previous works demand informative node features composed of user personal and behavioral information~\cite{dou2020enhancing,li2019spam}.
However, the privacy and data retention policies may restrict the access of user information by companies in practice.
Besides the challenge of constructing the graph, most real-world fraud detection tasks suffer from label scarcity issues since data annotation is labor-intensive due to the adversarial nature of fraudsters~\cite{dou2020enhancing}.


In this paper, we propose an approach that tackles two challenges above.
\textbf{1)} For the challenge of graph construction in practice, we propose a generic graph structure and node feature initialization approach.
We first introduce a graph transformation technique to convert commonly-used industrial graphs into smaller graphs while retaining useful information for downstream models.  
Then, inspired by recent work on node feature initialization for GNNs~\cite{duong2019node,cui2021positional}, we leverage graph topological features to initialize node features for the non-attributed graph.
\textbf{2)} To alleviate the label scarcity problem, we leverage the graph pre-training strategy, which is able to leverage more unlabeled data via graph contrastive learning~\cite{huyan2021unsupervised, tack2020csi,chen2021gccad}.
Specifically, we devise a self-supervised GNN pre-training framework to capture the graph's topological properties across the unlabeled data.
Then we generate inductive node embedding into the labeled dataset via the pre-trained graph encoder to train the fraud classifier. 

Our framework has been validated by experimental results on a large-scale industrial dataset. The proposed graph construction approach and graph pre-training strategy can improve learning efficiency and boost fraud detection performance.
Our contributions are summarized as follows:

\begin{itemize}
    \item We propose a graph construction method for GNN-based fraud detection on the non-attributed graph.
    
    \item A graph pre-training strategy is adopted to alleviate the label scarcity problem in the fraud detection problem.
    
    \item Experimental results on a large-scale real world dataset validate various combinations of approaches.
\end{itemize}




\section{Preliminary and Problem Definition}
\label{sec02-prelim}
We take the loan-default detection task to demonstrate the widely-used graph prototype for fraud detection.
As Fig.~\ref{fig:graph_structure} (a) shows, a personal loan dataset usually includes entities like address, device, loan, and user.
Most previous works build a \textit{multi-entity graph} composed of different entities as nodes and their relations as edges (e.g., user-has-loan, user-uses-device)~\cite{liu2018heterogeneous, li2019spam, liu2021intention}.
We can formally define the above non-attributed multi-entity graph as $\mathcal{G}_{m} = (\mathcal{V}_m, \mathcal{E}_m, \mathcal{O}_{\mathcal{V}}, \mathcal{R}_{\mathcal{E}})$, where ${v_i}\in \mathcal{V}_m$ denotes the nodes, $\mathcal{E}_m$ denotes the edges. $\mathcal{O}_{\mathcal{V}}$ ($\mathcal{R}_{\mathcal{E}}$ resp.) represents the node types (relation types resp.).

\noindent \textbf{Problem Definition.} Given a non-attributed multi-entity graph, we first transfer it into an attributed \textit{single-entity graph}
$\mathcal{G}_{s} = (\mathcal{V}_s,\mathcal{E}_s, \mathbf{X}^e, \mathbf{X}^v)$, where every ${v_i}\in \mathcal{V}_s$ belongs to the target entity to be classified.
$\mathbf{X}^e$ and $\mathbf{X}^v$ represent the edge feature matrix and node feature matrix, respectively.
With $\mathcal{G}_{s}$ and a set of partially annotated node labels $y_{i}\in Y, y_{i}\in(0,1)$, where 0 (1 resp.) represents the \textit{benign} (\textit{suspicious} resp.) entity,
we aim at training a classifier $f: \mathcal{G}_{s} \rightarrow Y $ to learn the representation of every ${v_i}\in \mathcal{V}_s$ and predict their labels.



\section{Methodology}
\label{sec03-method}

In this section, \textbf{1)} we first elucidate how we transform an attributed single-entity graph from a non-attributed multi-entity graph.
Then, \textbf{2)} we present a graph pre-training strategy with contrastive learning to leverage the unlabeled data.
Finally, \textbf{3)} we introduce how we encode the final node representation for fraud classification and how to fine-tune the GNN encoder. 


{\centering
\begin{figure}[!hbtp]
    \includegraphics[width=0.95\linewidth]{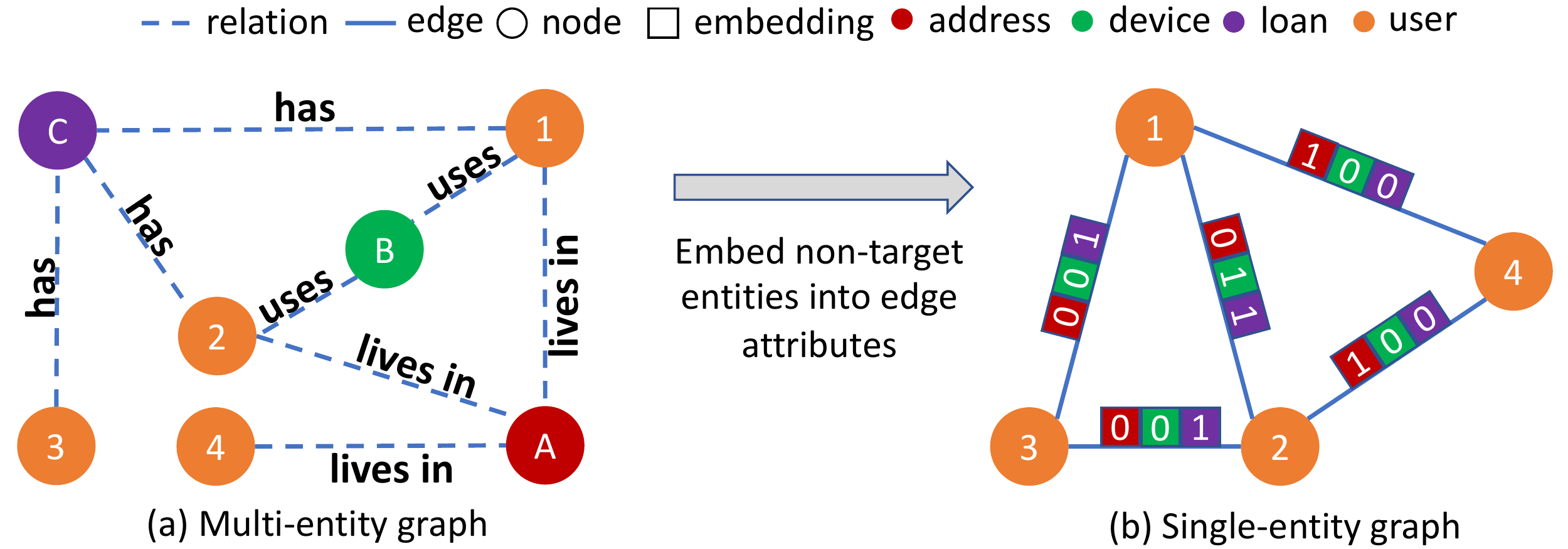}
    \caption{\small Transferring a multi-entity graph into a single-entity graph.
 }
\label{fig:graph_structure}
\end{figure}
}

\subsection{Graph Construction}

\subsubsection{Structure Transformation}
\label{sec03:structure}
As Fig.~\ref{fig:graph_structure} (a) showing, many industrial graphs may contain entities like address and device, which only play the role to connect similar target entities (\textit{i.e.}, users) while it is not easy to extract features from them.
Since GNNs' capability and efficiency are largely dependent on node feature~\cite{duong2019node} and graph size~\cite{zeng2019graphsaint}, respectively, we propose to transfer all non-target entities as the feature of edges between target entities (as shown in Fig.~\ref{fig:graph_structure}).
Specifically, the edge feature is a $d$-dimension one-hot vector, where $d$ is the number of all non-target entity types, and each dimension indicates whether two end target entities are connected via the corresponding non-target entity. 

The structure transformation above has three benefits: \textbf{1)} the single-entity graph shrinks the graph size significantly from the multi-entity graph via only keeping the target entities as nodes;
\textbf{2)} the connections and their importance between target entities and different non-target entities are retained and can facilitate GNNs;
\textbf{3)} for a center target node, a GNN could perceive more neighbor's information on the single-entity graph than the multi-entity graph with the same number of layers.

\subsubsection{Node Feature Initialization}
\label{sec03:feature}
GNNs aim at learning node representations by learning the similarities shared between connected nodes. 
However, the expressive ability of a GNN is highly dependent on the quality of node features \cite{duong2019node, cui2021positional}.
Therefore, given the single-entity graph without target entity features, it would be better to initialize node features before feeding the graph into GNNs. 
Considering the costly feature engineering and the adversarial nature of fraudsters, we resort to adopting the following four graph topological features for an expeditious node feature initialization~\cite{duong2019node,cui2021positional}.:
\begin{itemize}
    \item \textit{Random}: generating a feature vector for each node via sampling from a Gaussian distribution.
    \item \textit{Degree}~\cite{hamilton2017inductive}: converting the node degree into a one-hot degree vector for each node, where the vector dimension depends on the maximum degree across all nodes.
    \item \textit{PageRank}~\cite{brin1998anatomy}: computing the PageRank score for each node, and use it as the node feature.
    \item \textit{Eigen}~\cite{huang2020combining}: applying the eigen decomposition on the normalized adjacency matrix of $\mathcal{G}_{s}$ and the top-$k$ eigenvalues are the $k$-dimensional feature vector for each node.
\end{itemize}

We compare the performance of four nodes feature initialization approaches in Section~\ref{subsec: overall comparison}.



\subsection{Graph Pre-training}
{\centering
\begin{figure}[!hbtp]
    \includegraphics[width=\linewidth]{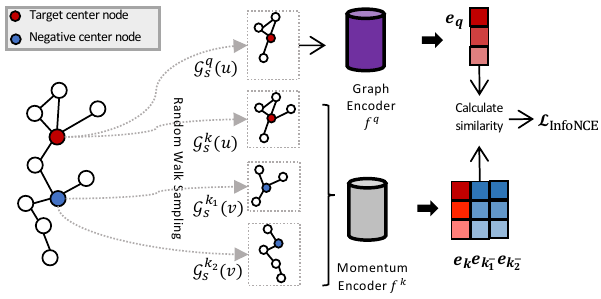}
    \caption{\small Pre-training the GNN encoder on the unlabeled data.}
\label{fig:graph_flowchart1}
\end{figure}
}

As we mentioned in Section~\ref{sec01-intro}, many real-world fraud detection tasks suffer from the label scarcity problem.
Though GNNs can leverage unlabeled data during training, their capability is limited within the local neighborhood of target entities.
Thanks to the recent advance in graph pre-training~\cite{qiu2020gcc, you2020graph}, we adopt self-supervised contrastive learning to encode more information from unlabeled data.
Specifically, under the assumption that common and transferable structural patterns exist across different nodes' sub-graphs, we can use plenty of unlabeled nodes to pre-train a GNN encoder to capture the local structure similarity between different nodes.
Our graph pre-training framework has three steps:
\textbf{1)} data augmentation, which constructs positive and negative sub-graph pairs of a node;
\textbf{2)} GNN encoder, which maps one node's structural pattern to the latent representation;
\textbf{3)} encoder training, which optimizes the GNN encoder with a contrastive loss. 

\subsubsection{Data augmentation}
Contrastive learning requires one instance's positive pairs and negative samples to enhance the classifier's discriminative capability.
Similar to previous work~\cite{you2020graph}, we construct positive pairs as two sub-graphs of a node.
Since two sub-graphs should share similar structure information to guarantee them to be positive pairs, we sample them from the $r$-ego network of a node.

We first conduct two iterations of random walk on node $u$'s $r$-ego network $\mathcal{G}_{s}(u)$ (the superscript is ignored for simplicity), to generate two sub-graphs $\mathcal{G}^{q}_{s}(u)$ and $\mathcal{G}^{k}_{s}(u)$, which are regarded as a positive pair.
After constructing positive sub-graph pairs for a node, we regard the sub-graphs generated from different nodes $v$'s $r$-ego networks as the negative samples of the node. Fig.~\ref{fig:graph_flowchart1} demonstrates the sub-graph construction process, where $\mathcal{G}^{q}_{s}(u)$ and $\mathcal{G}^{k}_{s}(u)$ are a positive pair since they are sampled from the same node.
$\mathcal{G}^{k_1}_{s}(v)$ and $\mathcal{G}^{k_2}_{s}(v)$ denote the negative samples of node $u$, which are sub-graphs sampled from the $r$-ego network of a different node.

\subsubsection{GNN encoder}
After retrieving positive and negative sub-graphs, we feed them into two graph encoders $f^{q}$ and $f^{k}$, which are illustrated in Figure~\ref{fig:graph_flowchart1}.
We encode the sub-graph $\mathcal{G}_{s}(u^{q})$ with graph encoder $f^q$, while encoding other sub-graphs with $f^k$.
Correspondingly, we generate low-dimensional representation vectors $\mathbf{e}_{q}$ and $\mathbf{e}_{k}$ for the positive pair $\mathcal{G}_{s}(u^{q})$ and $\mathcal{G}_{s}(u^{k})$, respectively.
We use the Graph Isomorphism Network (GIN)~\cite{xu2018powerful} as the GNN encoder to learn the node representation by only considering the node feature.
This approach has two advantages:
\textbf{1)} increasing the migration and generalization capabilities of the graph structure;
\textbf{2)} decoupling the correlation between node features and edge features.
The GNN encoder aggregation function is:
\begin{equation}
  \mathbf{x}_{i}^{v}{'}=\text{MLP}\Big((1+\epsilon)\cdot \mathbf{x}_{i}^v + \sum_{j\in \mathcal{N}(i)}\mathbf{x}_{j}^v\Big),  
\label{eq:2}
\end{equation}
where $\mathbf{x}_{i}^{v}{'}$ is $v_i$'s embedding at the next GNN layer, $\epsilon$ is the weight parameter, $\mathbf{x}_{i}^{v}$ is $v_i$'s embedding at the current layer, and $\mathbf{x}_{j}^{v}, j\in \mathcal{N}$ is the neighbor node embedding of $v_i$.

\subsubsection{Encoder training}

We adopt the contrastive loss named InfoNCE \cite{oord2018representation} to optimize the graph encoder in a self-supervised fashion, which maximizes the agreements between positive pairs. The InfoNCE loss is formulated as follows:

\begin{equation}
  \mathcal{L}_{\text{InfoNCE}}=-\log\frac{\text{exp}(\mathbf{e}_{q}^\intercal\mathbf{e}_{k}/\boldsymbol{\tau})}{\sum_{i=1}^{n} \text{exp}(\mathbf{e}_{q}^\intercal \mathbf{e}_{i}/ \boldsymbol{\tau})},
\label{eq:1}
\end{equation}
where $\boldsymbol{\tau}$ is the temperature hyper-parameter. 
Minimizing Eq.~(\ref{eq:1}) is equivalent to maximizing the similarity between positive pairs, \textit{i.e.}, $\mathbf{e}_{q}$ and $\mathbf{e}_{k}$, while minimizing the similarity between negative pairs, \textit{i.e.}, $\mathbf{e}_{q}$ and $\mathbf{e}_{i}$ where $i\neq k $.
In practice, we view those instances as a query embedding $\mathbf{e}_{q}$ and a set of key embeddings $\{\mathbf{e}_{i}\}|_{i=1}^{n}$.
The contrastive loss looks up a single key (denoted by $\mathbf{e}_{k}$) that $\mathbf{e}_{q}$ matches in
the key set. 

In contrastive learning, maintaining a K-size look-up key set is essential.
Intuitively, as the denominator in Eq.~(\ref{eq:1}) expresses, a larger key set size leads to better sampling of the underlying data space.
To further improve the optimization process, we adopt the MoCo~\cite{he2020momentum} training scheme, which maintains a dynamic set of keys with a queue and a moving-averaged encoder.
MoCo is able to increase the key set size without additional back-propagation costs.
Formally, if denoting the parameters of $f_k$ as $\theta_k$ and those of $f_q$ as $\theta_q$, MoCo updates $\theta_k$ as $\theta_k \leftarrow m\theta_k +(1-m)\theta_q$, where $m\in [0,1)$ is a momentum hyper-parameter.


\subsection{Model Fine-tuning}
The pre-trained GNN encoder is then employed on the labeled graph to fine-tune embeddings. Specifically, we sample a sub-graph for each node and fine-tune the pre-trained encoder to encode the sub-graph. The objective for this fine-tuning step is to predict the associated label of each node. To note, 
in addition to node features, 
we aggregate both node and edge features
of neighbors for final representation learning:
\begin{equation}
  \mathbf{x}_{i}^{v}{''}=\text{MLP}\Big((1+\epsilon)\cdot \mathbf{x}_{i}^v + \sum_{j\in \mathcal{N}(i)}\text{ReLU}(\mathbf{x}_{j}^v + \mathbf{x}_{ij}^e)\Big),  
\label{eq:3}
\end{equation}
where $\mathbf{x}_{ij}^e$ is the edge features between node $i$ and $j$. After passing the final node embedding to an MLP classifier, we adopt following cross-entropy loss to fine-tune the GNN encoder and the MLP parameters:

\begin{equation}\label{eq:ce_loss}
    \mathcal{L}_{\text{CE}} = \sum_{v_i\in \mathcal{V}_{train}}-\log\left(y_{i}\cdot\mbox{ReLU}(\mbox{MLP}(\mathbf{x}_{i}^{v}{''}))\right),
\end{equation}
where $\mathbf{x}_{i}^{v}{''}$ represents $v_i$'s embedding after GNN's last layer.


\section{Experiment}
\label{sec04-exp}
\subsection{Dataset}
\label{sec04:dataset}
We evaluate our method on a large-scale industrial dataset.
In Table~\ref{tab:data}, by comparing the statistics of the multi-entity graph $\mathcal{G}_{m}$ and the single-entity graph $\mathcal{G}_{s}$, there is a large discrepancy between them.
By transforming $\mathcal{G}_{s}$'s structure, the number of nodes has been reduced 26.69 times and edges by 22.07 times comparing with $\mathcal{G}_{m}$.
Like many real-world fraud detection cases, the labeled data is very small.
Only 0.188\% of nodes are labeled on the graph used in this paper, which contains 1482 fraudulent and 1287 benign target entities.
Utilizing the unlabeled data is critical for success, and our method has demonstrated the effectiveness of such a setting.


\begin{table}[!hbtp]
\centering
  \caption{\small Dataset Statistics.}
  \label{tab:data}
  \begin{tabular}{c|c|c|c|c}
  \hline
     & \verb|#| target entity & \verb|#| non-target entity &  $|\mathcal{O}_{\mathcal{V}}|$ & $|\mathcal{E}|$\\
    \hline
    $\mathcal{G}_{m}$ & 1,469,149 & 37,694,849 & 6 & 113,375,579\\
    \hline
    $\mathcal{G}_{s}$ & 1,469,149 & 0 & 1 & 5,136,750\\
    \hline
    $\mathcal{G}_{m}$:$\mathcal{G}_{s}$ & \multicolumn{2}{c|}{26.69} & 6 & 22.07\\
    \hline
  \end{tabular}
\end{table}

\subsection{Experimental Settings}
We compare four feature initialization methods proposed in Section~\ref{sec03:feature} under various settings. 
We adopt a 3-layer GNN encoder for all settings.
For pre-training, batch size is 200, embedding dimension is 16, and learning rate is $1e^{-6}$.
For fine-tuning, batch size is 100, embedding dimension is 32, and learning rate is $1e^{-5}$.
We conduct five-fold experiments and report the average \textit{micro-F1 score} in Table~\ref{tab:result}.

\subsection{Result Analysis}
\label{subsec: overall comparison}

\label{sec04:analysis}

\begin{table}[!hbtp]
\centering
  \caption{\small Fraud detection performance (micro-F1 score), ``PT'' represents using pre-training, ``NE'' and ``SE'' represent node embedding and sub-graph embedding, respectively.}
  \label{tab:result}
  \begin{tabular}{c|c|c|c|c}
  \hline
  \multirow {2}*{\textbf{Method}}
  &\multicolumn{2}{c|}{\textbf{multi-entity graph}}&\multicolumn{2}{|c}{\textbf{single-entity graph}}\\
  \cline{2-5}
    & NE & SE & NE & SE\\
    \hline
    Random &0.436&0.439& 0.441 & 0.447 \\
    Random + PT &0.424&0.427& 0.405 & 0.410 \\
    \hline
    Degree &0.472&0.481&0.488 & 0.479\\
    Degree + PT &0.499&0.483&0.532& 0.541\\
    \hline
    PageRank&0.566&0.559& 0.594 &0.610\\
    PageRank + PT &0.641&0.633& 0.661 &0.674\\
    \hline
    Eigen &0.630&0.628& 0.679 &0.683\\
    Eigen + PT &0.623 &0.644& 0.708 & \textbf{0.721}\\
    \hline
    \end{tabular}
\end{table}



Besides the node embedding, we also use the 3-hop sub-graph embedding for fraud classfication.
Table~\ref{tab:result} shows the experimental results, and we have the following findings:

\subsubsection{The single-entity graph is better than the multi-entity graph}
The best classification result in the single-entity graph exceeds that of the multi-entity graph by $11.95\%$.
For center node, a GNN can reach farther target entities on the single-entity graph compared to that of multi-entity graphs, thus more informative nodes are encoded on the single-entity graph.

\subsubsection{Graph pre-training is helpful}
We can observe a performance boost for most feature initialization methods after applying the proposed graph pre-training strategy (see method + PT in Table~\ref{tab:result}).
This result indicates the effectiveness of using the contrastive learning to pre-train the GNN encoder on the unlabeled data, and it boosts $5.56\%$ micro-F1 score compare Eigen + PT with Eigen.
On the contrary, graph pre-training has a negative effect on randomly initialized nod features.
A possible reason is that the random feature carries no topological information of the node and thus can not be leveraged by self-supervised contrastive learning.

\subsubsection{Sub-graph embedding is more effective than node embedding}
As shown in Table~\ref{tab:result}, using sub-graph embedding in the single-entity graph has noticeable improvement while the multi-entity graph is not.
It suggests that the single-entity graph has a better representation than the multi-entity graph with the same hops, and insufficient neighbor information will bring more harm rather than gain.

\subsection{Embedding Visualization}
{\centering
\begin{figure}[!hbtp]
    \includegraphics[width=\linewidth]{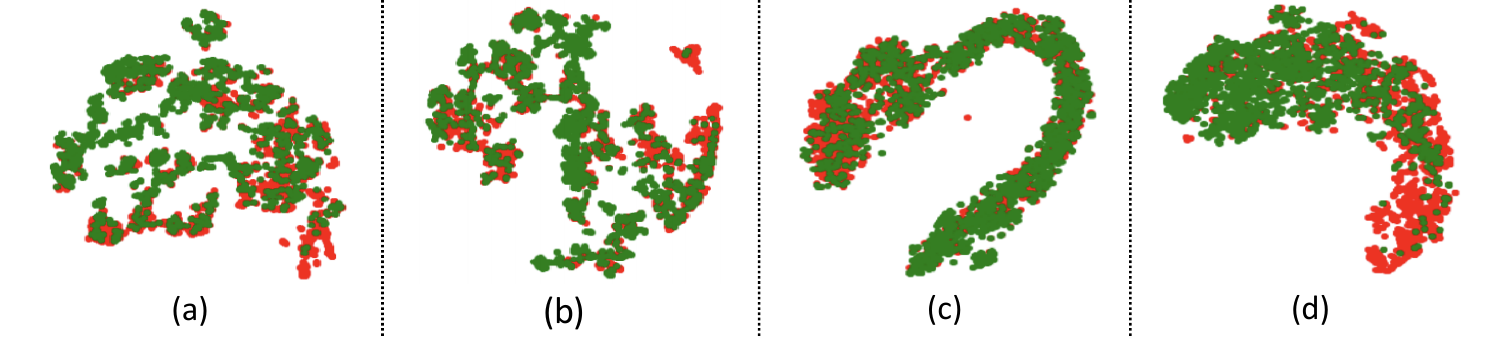}
    \caption{\small t-SNE plots of node embeddings obtained by: (a) Eigen + multi-entity; (b) Eigen + single-entity; (c) Eigen + PT + multi-entity; (d) Eigen + PT + single-entity. Red: fraudster, green: benign entity.
 }
\label{fig:tsne}
\end{figure}
}
To understand the difference between the proposed methods straightforwardly, we adopt t-SNE~\cite{liu2016visualizing} to visualize the node and sub-graph embeddings generated by the GNN encoders in Fig.~\ref{fig:tsne}.
It can be seen from the figure that graph pre-training not only gathers similar nodes together but also increases the discriminative capability of the GNN encoder.

\section{Conclusion and Future Work}
In this paper, we propose a framework to transform a general non-attributed graph in industry and initialize its node and edge features for GNN-based fraud detection.
The experimental results on a large-scale industrial dataset demonstrate the effectiveness of the proposed framework.
Future work includes investigating more informative node feature initialization approaches and optimize the graph pre-training efficiency.



\bibliographystyle{IEEEtran}
\bibliography{bigdata}

\end{document}